\documentclass[
]{style/ceurart}

\usepackage{listings}
\usepackage{tcolorbox}
\usepackage[ruled,vlined]{algorithm2e}

\begin{document}

\copyrightyear{2025}
\copyrightclause{Copyright for this paper by its authors.
  Use permitted under Creative Commons License Attribution 4.0
  International (CC BY 4.0).}

\conference{LaCATODA 2026: The 10th Linguistic and Cognitive Approaches to Dialog Agents Workshop at the 40th AAAI conference,
  January 20--27, 2026, Singapore Expo}

\title{CID-GraphRAG: Enhancing Multi-Turn Dialogue Systems through Dual-Pathway Retrieval of Conversation Flow and Context Semantics}


\author[1]{Ziqi Zhu}[%
email=ziqizhu@amazon.com,
]

\author[2]{Tao Hu}[%
email=terrence.hu@huolala.cn,
]

\author[2]{Honglong Zhang}[%
email=damon2.zhang@huolala.cn,
]

\author[2]{Dan Yang}[%
email=dandan.yang@huolala.cn,
]

\author[2]{Hangeng Chen}[%
email=hanke.chen@huolala.cn,
]

\author[2]{Mengran Zhang}[%
email=zmrmira.zhang@huolala.cn,
]

\author[3]{Xilun Chen}[%
email=123090042@link.cuhk.edu.cn,
]

\address[1]{Amazon Web Services}
\address[2]{HuoLaLa}
\address[3]{The Chinese University of Hong Kong, Shenzhen}

\begin{abstract}
We present CID-GraphRAG (\textbf{C}onversational \textbf{I}ntent-\textbf{D}riven \textbf{Graph} \textbf{R}etrieval-\textbf{A}ugmented \textbf{G}eneration), a novel framework that addresses the limitations of existing dialogue systems in maintaining both contextual coherence and goal-oriented progression in multi-turn customer service conversations. Unlike traditional RAG systems that rely solely on semantic similarity or static knowledge graphs, CID-GraphRAG constructs intent transition graphs from goal-achieved historical dialogues and implements a dual-retrieval mechanism that balances intent-based graph traversal with semantic search. This approach enables the system to simultaneously leverage both conversational intent flow patterns and contextual semantics, significantly improving retrieval quality and response quality.
In extensive experiments on real-world customer service dialogues, we demonstrated that CID-GraphRAG significantly outperforms both semantic-based and intent-based baselines across automatic metrics, LLM-as-a-Judge evaluations and human evaluations, with relative gains of $11.4\%$ in BLEU, $4.9\%$ in ROUGE, and $5.9\%$ in METEOR. Most notably, CID-GraphRAG achieves a $57.9\%$ improvement in response quality according to LLM-as-a-Judge evaluations. These results demonstrate that integrating intent transition structures with semantic retrieval creates a synergistic effect that neither approach achieves independently, establishing CID-GraphRAG as an effective framework for real-world multi-turn dialogue systems in customer service and other knowledge-intensive domains.

\end{abstract}


\maketitle

\section{Introduction}
Large Language Models (LLMs) face significant challenges in domain-specific multi-turn dialogues despite Retrieval-Augmented Generation (RAG) enhancements \cite{Lewis2020}. Customer service dialogues present unique challenges: temporal dependencies creating complex reference chains across turns, and dynamically evolving conversation objectives that traditional RAG approaches fail to address effectively.

Current research follows two separate trajectories - Conversation RAG methodologies \cite{roy2024learning,ye2024boosting} optimize dialogue history representation through semantic matching, while GraphRAG approaches \cite{graphrag2025,graphrag2024,zhang2025credible} leverage knowledge graphs for reasoning over static domain knowledge. Both have fundamental limitations—Conversation RAG lacks structural guidance for conversation flow, while GraphRAG fails to model intent transitions essential for coherent multi-turn dialogues.

We propose CID-GraphRAG (\textbf{C}onversational \textbf{I}ntent-\textbf{D}riven \textbf{Graph} \textbf{R}etrieval-\textbf{A}ugmented \textbf{G}eneration), a novel framework that integrates intent transition graphs with semantic similarity retrieval. 
Our approach leverages goal-achieved historical conversations to model conversation dynamics through both semantic content and conversation flow, enabling more coherent and goal-oriented responses. The key contributions of our work include:
\begin{itemize}
\item Intent transition graph construction: automatically building intent transition graphs from goal-achieved historical dialogues, systematically cataloging hierarchical intents from both assistant and user utterances to create a network of effective conversational trajectories.
\item Dual-pathway retrieval: balancing intent transition graph with semantic similarity search, through a weighting mechanism. This integration enables the system to simultaneously leverage conversation structure and contextual semantics, addressing the limitations of single-signal approaches in complex multi-turn dialogues.
\end{itemize}
Experiments on real-world customer service dialogues demonstrate that CID-GraphRAG significantly outperforms both semantic-only retrieval and intent-only graph approaches across all evaluation metrics, highlighting the effectiveness of our integrated dual-pathway approach.

\section{Related Works}
RAG enhances Large Language Models (LLMs) by integrating external knowledge sources, improving response accuracy and informativeness in dialogue systems \cite{Lewis2020}. Recent advancements include memory-based approaches like DH-RAG \cite{zhang2025dhrag} and SELF-multi-RAG \cite{self-rag}, evaluation frameworks like MTRAG \cite{katsis2025mtrag}, and specialized applications combining knowledge graphs for customer service \cite{xu2024retrieval} and real-time problem detection with FAQ retrieval \cite{agrawal2024beyond}.

GraphRAG enhances traditional RAG by integrating knowledge graphs for structured semantic retrieval and multi-hop reasoning \cite{graphrag2025}, better capturing contextual dependencies in multi-turn dialogues \cite{graphrag2024}. Key advantages include enhanced retrieval accuracy through preserved structural relationships \cite{xu2024retrieval}, effective processing of multi-hop queries requiring cross-fragment reasoning \cite{graphrag2025,zhang2025credible}, context continuity maintenance via entity relationship tracking \cite{xu2024retrieval}, and improved answer interpretability through visualized reasoning paths \cite{linders2025knowledge}.

Conversation RAG addresses multi-turn dialogue challenges in customer service, where user queries often depend on historical interactions \cite{roy2024learning,ye2024boosting}. Key advancements include context-aware frameworks like DH-RAG \cite{zhang2025dhrag} and SELF-multi-RAG \cite{self-rag}; structured retrieval approaches such as KG-RAG \cite{xu2024retrieval}; query decomposition techniques like Collab-RA \cite{collab-rag}; and specialized evaluation benchmarks including RAD-Bench \cite{kuo2024radbench}, MTRAG \cite{katsis2025mtrag}, and ChatQA \cite{chatqa}. While these approaches have advanced independently, no unified solution simultaneously addresses both contextual coherence and structured reasoning challenges. 

\begin{figure}[t]
\centering
\includegraphics[width=1.0\linewidth]{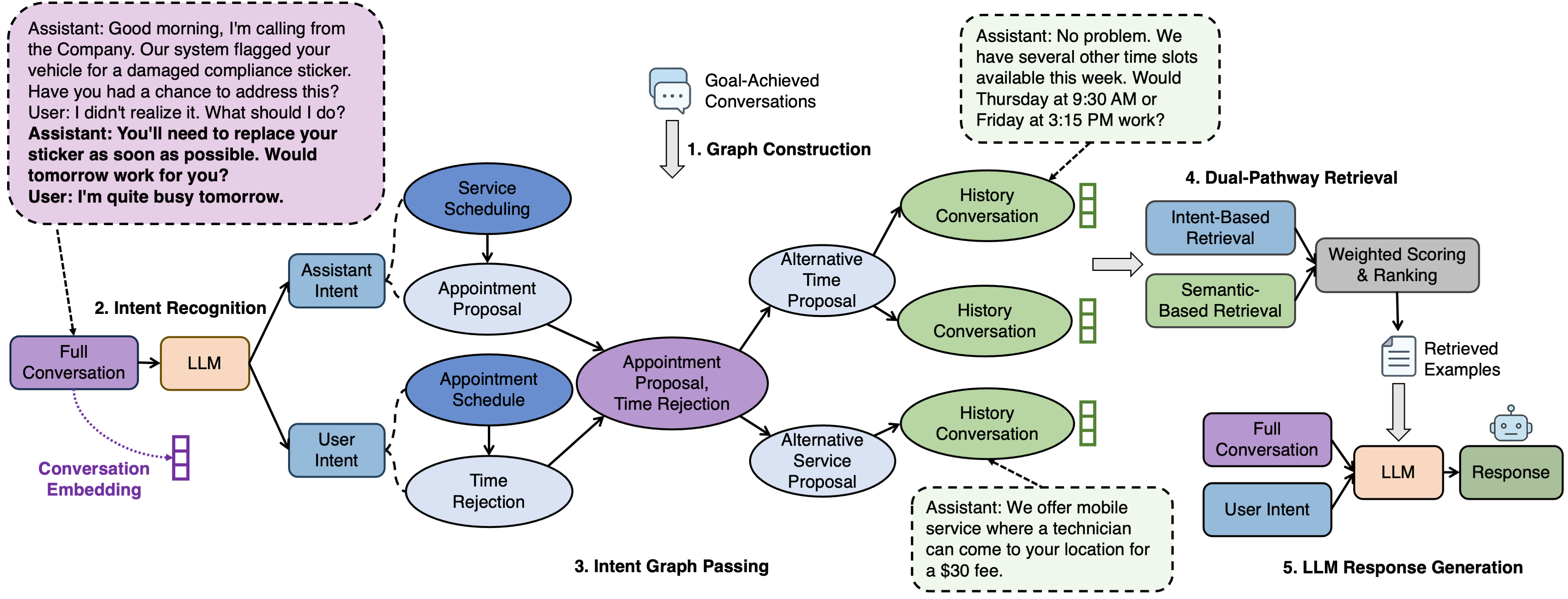}
\caption{The detailed framework of CID-GraphRAG. The CID-GraphRAG consists of two phases: (1) a construction phase that builds an intent graph from goal-achieved conversations, and (2) an inference phase that identifies user and assistant intents from current dialogue turn, retrieves high-quality examples from both intent-based and semantic-based pathway via weighting mechanism, and uses LLM for structured response generation.}
\label{fig3.1}
\end{figure}

\section{Methodology}

CID-GraphRAG innovatively integrates Conversation RAG and GraphRAG to address the challenge of simultaneously maintaining contextual coherence and goal-oriented progression in customer service dialogues. Our methodology consists of three key components: intent graph construction, dual-pathway retrieval, and structured response generation.

\subsection {System Overview}  
CID-GraphRAG operates in two complementary phases (Figure \ref{fig3.1}).

\begin {itemize}
\item Construction Phase: We leverage LLM to perform intent recognition on goal-achieved historical conversations and build an intent graph that captures both conversation structure and semantic content.
\item Inference Phase: We identify user and assistant intents of current dialogue turn, then employ dual-pathway retrieval that balances intent patterns with semantic similarity to identify relevant examples for response generation.
\end {itemize}

\subsection {Construction Phase}  
To effectively model conversation dynamics, we first need a structured representation of dialogue intents and their relationships. This process involves two key components: intent recognition and graph construction.

\subsubsection {Intent Recognition}  
We propose a hierarchical dual-layer intent framework that systematically categorizes conversational actions at different levels of granularity:

\begin{itemize}
    \item \textbf{Primary intents ($I_{1}$)} represent broad functional categories of dialogue turns, capturing the general conversational purpose (e.g., "Service Scheduling"),
    \item \textbf{Secondary intents ($I_{2}$)} encode fine-grained conversational actions within each primary category, specifying the precise communicative function (e.g., "Appointment Proposal").
\end{itemize}

Our intent recognition process employs a two-stage classification approach powered by LLM, first identifying primary intents, then identifying secondary intents within that category. This hierarchical approach improves intent classification accuracy compared to direct fine-grained classification. Appendix \ref{app:intent_recognition} provides concrete examples demonstrating how this framework disambiguates utterances with identical wording but different contextual meanings.

The complete classification procedure is formalized in Algorithm \ref{alg:intent_classification}, where we use Claude 3.7 Sonnet to classify both assistant and user utterances in each dialogue turn.

\subsubsection {Graph Construction} 
\begin{figure}[t]
\centering
\includegraphics[width=0.82\linewidth]{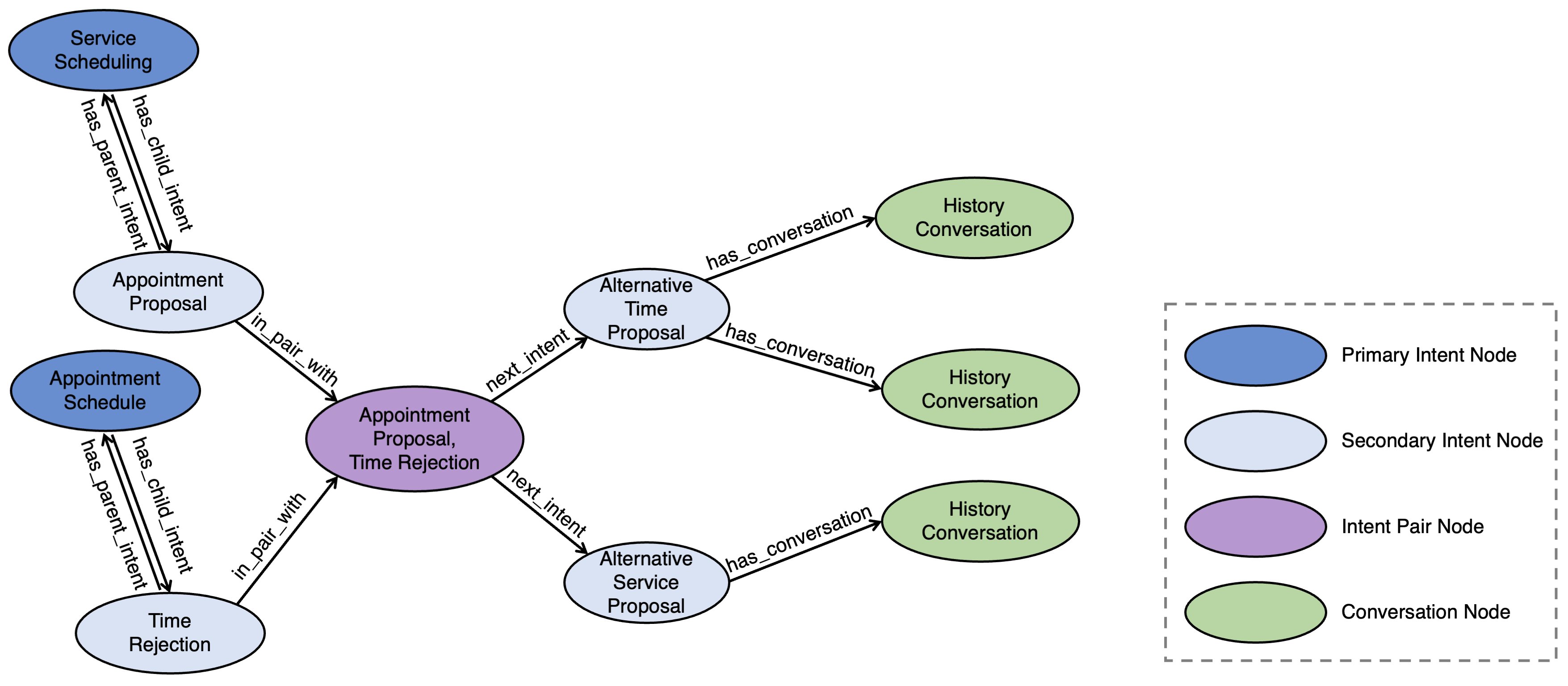}
\caption{Internal structure of the CID-Graph. The graph comprises distinct primary intent nodes, secondary intent nodes and conversation nodes. Key relations include hierarchical, pairing, transition and dialogue anchoring.}
\label{fig3.2}
\end{figure}

After intent recognition, we organize intents into a graph structure that captures conversation dynamics, as illustrated in Figure \ref{fig3.2}.
Our intent graph contains three node types:
\begin{itemize}
    \item \textbf{Intent nodes}: individual user and assistant intents. Denoted as $I_{1}^\text{assistant}, I_{2}^\text{assistant}, I_{1}^\text{user}$, and $I_{2}^\text{user}$.
    \item \textbf{Intent pair nodes}: combined assistant-user intent pairs. Denoted as $P(I_{2}^\text{assistant},I_{2}^\text{user})$.
    \item \textbf{Conversation nodes}: complete dialogue examples exhibiting specific intents. Denoted as $D_\text{hist}$.
\end{itemize}

    

\begin{algorithm}[H]
\SetAlgoLined
\DontPrintSemicolon
\caption{Two-Stage Intent Classification for Dialogue Turns}
\small
\label{alg:intent_classification}

\KwIn{Dialogue turn $T$ with current utterances and dialogue history}
\KwOut{Assistant intents $(I^{\text{assistant}}_1, I^{\text{assistant}}_2)$, User intents $(I^{\text{user}}_1, I^{\text{user}}_2)$}

\ForEach{\text{participant} $P \in \{\text{assistant}, \text{user}\}$}{
    \textbf{Stage 1:} Primary Intent Classification\;
    
    Construct prompt with:\;
    \Indp
        - Domain context\;
        - Dialogue history\;
        - Current dialogue turn\;
        - List of primary intents\;
    \Indm
    
    Send prompt to Claude 3.7 Sonnet\;
    Extract primary intent $I^P_1$ from response\;
    
    \vspace{0.5em}
    
    \textbf{Stage 2:} Secondary Intent Classification\;
    
    Construct prompt with:\;
    \Indp
        - Domain context\;
        - Dialogue history\;
        - Current dialogue turn\;
        - Primary intent $I_1^P$ determined in Stage 1\;
        - Candidate secondary intents specific to $I_1^P$\;
    \Indm
    
    Send prompt to Claude 3.7 Sonnet\;
    Extract secondary intent $I^P_2$ from response\;
}

\Return{$(I^{\text{assistant}}_1, I^{\text{assistant}}_2, I^{\text{user}}_1, I^{\text{user}}_2)$}
\end{algorithm}

These nodes are connected through hierarchical relations ($has\_child\_intent$, $has\_parent\_intent$), pairing relations ($in\_pair\_with$), transition relations ($next\_intent$), and dialogue anchoring ($has\_conversation$).
This graph structure enables the system to navigate conversation trajectories that have historically led to successful outcomes.

\subsection {Inference Phase}  
During the inference phase, we propose a dual-layer retrieval mechanism that integrates intent transition retrieval and semantic similarity retrieval. This dual-layer approach enables mutually-assisted retrieval, enhancing both retrieval effectiveness and efficiency.

\subsubsection{Dual-Pathway Retrieval}
\label{sec:dual_retrieval}
The core innovation of CID-GraphRAG is its dual-pathway retrieval mechanism that balances intent structure with semantic context. By simultaneously leveraging conversation flow patterns from successful historical dialogues and contextual semantics from the current interaction, this approach enables mutually-assisted retrieval that enhances both retrieval relevance and goal orientation.

\textbf{Intent-Based Retrieval Path: }The intent path captures structural patterns by:
\begin {enumerate}
\item Intent recognition: identifying the user's current intent $I_{2}^\text{user}$,
\item Intent pair generation: combining user intent with the previous assistant intent to form an intent pair $P(I_{2}^\text{assistant},I_{2}^\text{user})$,
\item Graph retrieval: retrieving candidate intents based on the normalized co-occurrence frequencies. To ensure compatibility with semantic similarity scores, we normalize frequencies using:
\begin{align}
f'(I_{2}^\text{candidate}) = \frac{f(I_{2}^\text{candidate})}{\max(f(I_{2}^\text{candidate}))}
\label{eq:normalized_freq}
\end{align}
\end {enumerate} where $f(I_2^{\text{candidate}})$ represents the co-occurrence frequency of candidate assistant intents. When encountering new intent combinations, a fallback mechanism identifies the most similar known patterns (detailed in Appendix \ref{app:fallback}).

\textbf{Semantic-Based Retrieval Path:} We integrate semantic retrieval within the intent graph (details in Appendix \ref{app:semantic_match}):
\begin{enumerate}
    \item Embeds the current dialogue history $D_\text{curr}$,
    \item Embeds historical dialogue examples $D_\text{hist}$ linked to candidate intents $I_2^{\text{candidate}}$,
    \item Computes cosine similarity $\text{sim}(D_\text{curr},D_\text{hist})$ to identify contextually relevant examples.
\end{enumerate}

\subsubsection {Weighted Integration}
For each candidate intent $I_2^{\text{candidate}}$, we retrieve associated historical dialogues $D_{\text{hist}}$. We then compute a weighted score $S_i$ for each dialogue $D_{\text{hist}}^i$ by combining intent-based and semantic-based signals:

\begin{equation}
\begin{split}
S_i & = \alpha \cdot f'(I_2^{\text{candidate}}) + (1-\alpha) \cdot \text{sim}(D_{\text{curr}}, D_{\text{hist}}^i)
\end{split}
\label{eq:weighted_score}
\end{equation} where $\alpha$ denotes a weight parameter that balances intent-based and semantic-based retrieval.

The system then selects the top-$k$ historical dialogues with the highest combined scores:
\begin{align}
\text{Top-}k\ D_{\text{hist}} = \arg\max_{i} (S_i) \label{eq:top_k_selection}
\end{align}

\noindent These top-k dialogues serve as few-shot examples for response generation.

\subsubsection {Response Generation}  

The response generation leverages the retrieved examples in a structured prompt containing:
\begin {itemize}
\item The dialogue history and current user query
\item The identified user intent
\item Top-k examples of similar dialogues with their assistant intents
\item Generation instructions
\end {itemize}
This approach grounds the LLM's response in both the conversation's structural patterns and its semantic context, enabling responses that maintain both coherence and goal orientation.

\section{Experiments}

\subsection {Experimental Setup}  

\subsubsection {Dataset}  
We conducted experiments on a real-world customer service dataset consisting of 268 conversations between representatives and drivers regarding vehicle sticker issues. The dataset exhibits strong task-orientedness and multi-turn complexity, covering multiple subtopics including compliance consultation, violation inquiries, appointment scheduling, service coordination, etc. All dialogues were manually selected and annotated by domain experts with extensive experience in customer service. A dataset example is shown in Appendix \ref{app:dataset_example}.

The dataset contains 1,574 total dialogue turns with an average of 5.9 turns per dialogue (ranging from 2 to 24 turns) and 22.0 words per utterance on average. All dialogues have been rigorously anonymized to protect user privacy. Following standard practice, we partitioned the dataset into training set ($80\%$, 214 dialogues with 1,299 conversation turns), validation set ($10\%$, 27 dialogues with 149 turns), and test set ($10\%$, 27 dialogues with 126 turns). 

\subsubsection {Evaluation Metrics}
\label{sec:eval_metrics}

We employed three complementary evaluation approaches:

\textbf{Automatic Metrics}: We calculated BLEU, ROUGE, METEOR and BERTSCORE between system response and ground truth responses, as well as between retrieved responses and ground truth responses.

\textbf{LLM-as-a-Judge Evaluation}: We utilized Claude 3.7 Sonnet \citep{sonnet_aws} as an independent evaluator to assess both retrieved response quality and generated responses. For each evaluation instance, the LLM was provided with the complete dialogue history, the current user query, anonymized system responses generated by different systems, and the ground truth response for reference. For each comparison, LLM compared outputs from all systems across five dimensions (relevance, helpfulness, style consistency, contextual appropriateness, and professionalism) and provided numerical ratings (1-10) for each system, with the highest-rated system designated as "winner" for that example. When multiple systems receive identical highest ratings, all are counted as winners. We employ both average ratings and win counts to analyze system performance. For retrieved quality evaluation, we excluded the "professionalism" dimension, and removed ground truth responses in the LLM prompt. The complete prompt template for response generation is provided in Appendix \ref{judge_prompt}.

\textbf{Human Evaluation}: We conducted a human evaluation on response generation with five domain experts experienced in customer service. The human evaluators performed blind assessments of responses generated by all systems across the same five dimensions with ratings (1-10). Evaluators were presented with the full conversation history, current user query, and anonymized system responses. This evaluation provides critical insights into real-world utility and performance aspects that may not be fully captured by computational methods.

\subsubsection {Baselines and Implementation}  
We compare CID-GraphRAG with three baselines:
\begin{itemize}
    \item \textbf{Direct LLM} generates responses using conversation history, without any retrieval augmentation;
    \item \textbf{Intent RAG} retrieves examples solely via intent graph matching, using random dialogue examples from the most frequent candidate assistant intent;
    \item \textbf{Conversation RAG} retrieves based on semantic similarity of conversation history.
\end{itemize}

All experiments used Claude 3.7 Sonnet (temperature=0.0) with 5-shot prompting for RAG methods and zero-shot for Direct LLM.

\subsection {Hyperparameter Analysis} 
We examined five different weight configurations for CID-GraphRAG on validation set to determine the optimal balance between intent-based and semantic-based retrieval. Table \ref{tab:weight_configs} presents the results.

\begin{table*}[t]
\caption{Automatic metric results in retrieval performance across all weight configurations. \textbf{Bold} values represent best performance for each metric.}
\label{tab:weight_configs}
\small
\centering
\renewcommand{\arraystretch}{1.2}
\begin{tabular}{lccccccc}
\hline
 &
\textbf{BLEU-2} &
\textbf{BLEU-4} &
\textbf{ROUGE-1} &
\textbf{ROUGE-2} &
\textbf{ROUGE-L} &
\textbf{METEOR} &
\textbf{BERTSCORE} \\ 
\hline
$\alpha$=0.1 &
\textbf{7.48} &
\textbf{2.89} &
\textbf{23.89} &
\textbf{5.78} &
\textbf{20.14} &
\textbf{14.76} &
\textbf{63.32} \\
$\alpha$=0.3 &
7.01 &
2.71 &
23.37 &
5.33 &
19.81 &
14.24 &
63.15 \\
$\alpha$=0.5 &
6.95 &
2.70 &
23.27 &
5.30 &
19.76 &
14.18 &
63.11 \\
$\alpha$=0.7 &
6.95 &
2.70 &
23.27 &
5.30 &
19.76 &
14.18 &
63.11 \\
$\alpha$=0.9 &
6.95 &
2.70 &
23.27 &
5.30 &
19.76 &
14.18 &
63.11 \\ 
\hline
\end{tabular}
\end{table*}

Our empirical analysis reveals two key findings regarding the parameter $\alpha$: (1) the configuration with $\alpha=0.1$ consistently achieves optimal performance across all evaluation metrics, with performance degradation as intent weight increases; and (2) the configurations with $\alpha \geq 0.5$ yield identical retrieval results, suggesting a saturation threshold beyond which intent-based retrieval dominates the ranking process. To validate these automatic evaluation results, we performed LLM-as-a-Judge evaluation using the same criteria from Section \ref{sec:eval_metrics}. Among the cases where retrieval results differed, $\alpha=0.1$ configuration achieved superior performance in 82.9\% cases. This alignment between automatic metrics and LLM-as-a-Judge evaluation validates our parameter selection approach.

These observations suggest two important insights regarding the dual-pathway retrieval mechanism. First, intent information provides a valuable structural signal that complements semantic retrieval by disambiguating between utterances with high lexical similarity but different conversational functions. Second, excessive weighting of intent information proves detrimental to system performance, as it prioritizes conversational patterns over semantic content, retrieving examples that maintain dialogue structure but potentially lack context relevance.

The optimal balance appears to be a predominantly semantic-driven approach ($90\%$) guided by a small but significant intent component ($10\%$) that helps the system maintain awareness of the conversation's structural dynamics. Based on these findings, we adopt $\alpha=0.1$ settings for our main experiments.

\subsection {Evaluation Results}  
\begin{table}[t]
\centering
\caption{Performance comparison across all metrics. \textbf{Bold} values indicates best performance for each metric. The percentage values in parentheses show CID-GraphRAG's relative improvement over the second-best system.}
\begin{tabular}{@{}l@{\hspace{3pt}}@{\hspace{3pt}}ccc@{\hspace{3pt}}@{\hspace{3pt}}cccc@{}}
\toprule
\multirow{3}{*}{} & \multicolumn{3}{c}{\textbf{Retrieval Quality}} & \multicolumn{4}{c}{\textbf{Response Generation}} \\
\cmidrule(lr){2-4} \cmidrule(lr){5-8}
& \textbf{Intent} & \textbf{Conv.} & \textbf{CID-} & \textbf{Direct} & \textbf{Intent} & \textbf{Conv.} & \textbf{CID-} \\
& \textbf{RAG} & \textbf{RAG} & \textbf{GraphRAG} & \textbf{LLM} & \textbf{RAG} & \textbf{RAG} & \textbf{GraphRAG} \\
\midrule
\multicolumn{8}{l}{\textbf{Automatic Metrics}} \\
\midrule
BLEU-4 & 2.26 & 3.40 & \textbf{3.50} (2.9\%) & 1.46 & 1.56 & 1.85 & \textbf{2.06} (11.4\%) \\
ROUGE-L & 19.08 & 21.06 & \textbf{22.15} (5.2\%) & 18.25 & 19.66 & 20.03 & \textbf{21.01} (4.9\%) \\
METEOR & 15.31 & 17.02 & \textbf{18.21} (7.0\%) & 18.13 & 20.48 & 21.32 & \textbf{22.58} (5.9\%) \\
BERTSCORE & 63.10 & 63.74 & \textbf{64.70} (1.5\%) & 62.54 & 63.32 & 63.91 & \textbf{64.41} (0.8\%) \\
\midrule
\multicolumn{8}{l}{\textbf{LLM-as-a-Judge Evaluation}} \\
\midrule
Style Consistency & 6.61 & 7.71 & \textbf{8.05} (4.4\%) & 7.83 & 7.48 & 7.82 & \textbf{8.21} (4.9\%) \\
Helpfulness & 4.94 & 7.09 & \textbf{7.75} (9.3\%) & 7.71 & 6.94 & 7.49 & \textbf{7.81} (1.3\%) \\
Context Approp. & 5.17 & 7.10 & \textbf{7.66} (7.9\%) & 7.63 & 7.40 & 7.74 & \textbf{7.81} (0.9\%) \\
Relevance & 5.39 & 6.88 & \textbf{7.74} (12.5\%) & 7.81 & 6.90 & 7.54 & \textbf{7.98} (2.2\%) \\
Professionalism & - & - & - & 7.68 & 7.19 & 7.63 & \textbf{8.01} (4.3\%) \\
Average & 5.53 & 7.20 & \textbf{7.80} (8.4\%) & 7.73 & 7.18 & 7.64 & \textbf{7.96} (3.0\%) \\
\midrule
\multicolumn{8}{l}{\textbf{Human Evaluation}} \\
\midrule
Style Consistency & - & - & - & 8.85 & 9.38 & 8.91 & \textbf{9.62} (2.6\%) \\
Professionalism & - & - & - & 8.59 & 9.03 & 8.40 & \textbf{9.12} (1.0\%) \\
Relevance & - & - & - & 7.94 & 7.63 & 7.56 & \textbf{8.28} (4.3\%) \\
Helpfulness & - & - & - & 9.00 & 9.01 & 9.07 & \textbf{9.22} (1.7\%) \\
Context Approp. & - & - & - & 9.19 & 9.07 & 9.09 & \textbf{9.40} (2.3\%) \\
Average & - & - & - & 8.71 & 8.83 & 8.61 & \textbf{9.13} (3.4\%) \\
\bottomrule
\end{tabular}
\label{tab:comprehensive_results}
\end{table}

\begin{figure}[h]
\centering
\includegraphics[width=0.8\linewidth]{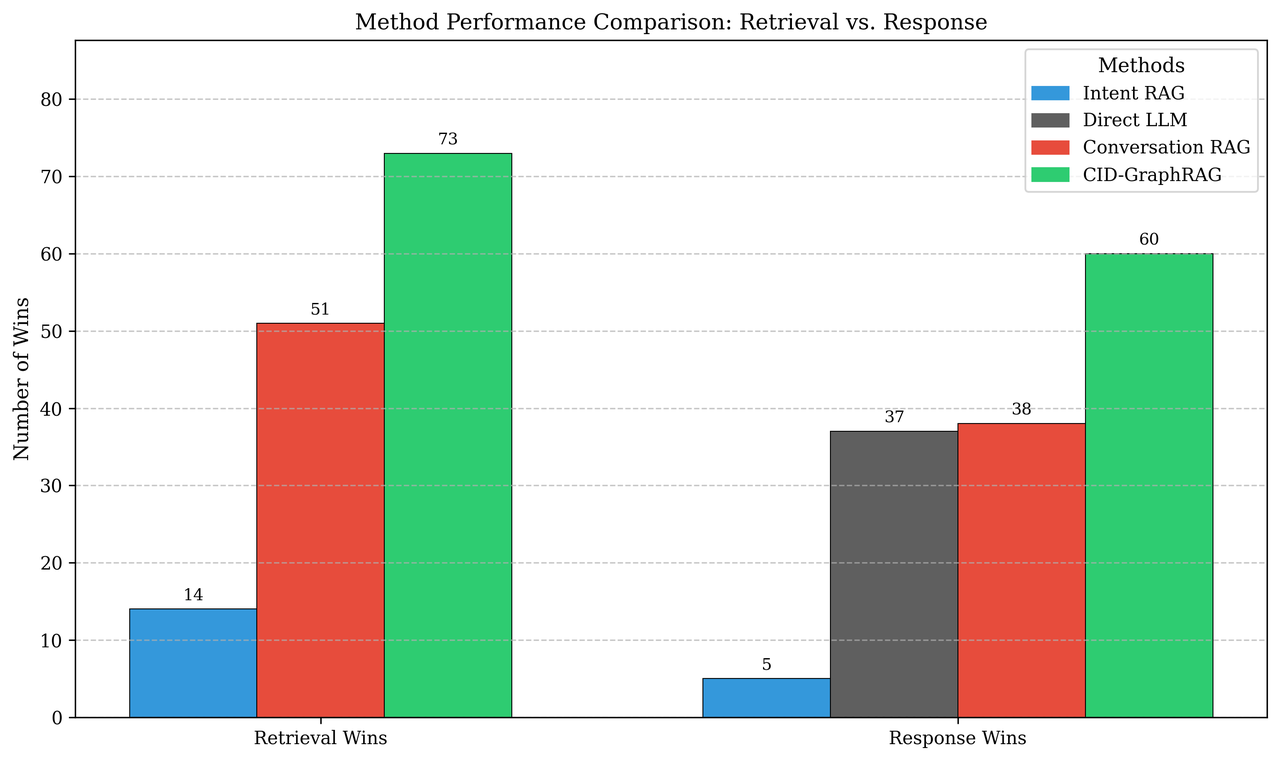}
\caption{Performance comparison of different methods based on LLM-as-a-Judge win counts. The left group shows retrieval quality wins, and the right group shows response generation wins. CID-GraphRAG consistently outperforms all baseline methods in both aspects, with particularly significant advantages in retrieval quality.}
\label{fig:win_counts}
\end{figure}

Table \ref{tab:comprehensive_results} presents comprehensive results comparing CID-GraphRAG with baselines for retrieval quality and response generation respectively. To better illustrate the overall performance of different systems in our comparative evaluation, Figure \ref{fig:win_counts} presents the win counts from LLM-as-a-Judge evaluations for both retrieval quality and response generation quality.

\textbf{Retrieval Quality:} CID-GraphRAG demonstrates consistent superiority in retrieval quality across all metrics. For automatic metrics, it achieves relative gains of 2.9\% in BLEU-4, 5.2\% in ROUGE-L, and a substantial 7.0\% in METEOR compared to the second-best system. The LLM-as-a-Judge evaluation reveals even more significant improvements, with CID-GraphRAG outperforming other approaches across all dimensions, particularly in relevance (+12.5\%) and helpfulness (+9.3\%), demonstrating that our dual-pathway approach retrieves more useful examples. These improvements are further validated by win count analysis (Figure \ref{fig:win_counts}), where CID-GraphRAG (73 wins) outperforms the second-best system Conversation RAG (51 wins) by 43\%.

\textbf{Response Quality:} The superior retrieval quality of CID-GraphRAG directly translates to improved response generation. Automatic metrics show consistent improvements, with CID-GraphRAG outperforming all baselines across BLEU, ROUGE, METEOR, and BERTSCORE. The most substantial improvement is observed in BLEU-4, with an 11.4\% gain over the second best.

LLM-as-a-Judge evaluation of generated responses reveals significant advantages. CID-GraphRAG achieves the highest scores across all dimensions, with notable improvements in style consistency (+4.9\%) and professionalism (+4.3\%). Most importantly, CID-GraphRAG records the highest number of wins (60 wins) in head-to-head comparisons, representing a 57.9\% improvement over the second best (Conversation RAG, 38 wins).

Human evaluation of response generation further validates these findings, with CID-GraphRAG achieving the highest average score (9.13) across all systems. The model shows particular strength in relevance (+4.3\%) and style consistency (+2.6\%), suggesting that the intent-driven approach produces responses that are both contextually appropriate and goal-oriented.

\subsection {Discussion}  

Our experimental results consistently demonstrate the superiority of CID-GraphRAG over other conventional RAG approaches. Despite allocating only a $10\%$ weight to intent information, CID-GraphRAG showed a substantial improvement over Conversation RAG in both retrieval quality and response generation. Intent information acts as a powerful guiding mechanism that helps filter out semantically similar but contextually inappropriate examples, reducing semantic drift in the retrieval process. 

Comparing human and LLM-as-a-Judge evaluations reveals interesting differences in system rankings. While both confirm CID-GraphRAG's superiority, LLM-as-a-Judge identified Direct LLM as competitive in style consistency (7.83) and relevance (7.81), whereas human evaluators consistently ranked Intent RAG second, particularly for style consistency (9.38) and professionalism (9.03). This suggests domain experts place higher value on intent-guided responses than LLM evaluators, especially for maintaining professional communication standards in customer service contexts.

\begin{table}[ht]
\centering
\caption{Computational efficiency comparison across different approaches}
\label{tab:performance_efficiency}
\begin{tabular}{lccccc}
\hline
\textbf{Approach} & \multicolumn{2}{c}{\textbf{Input Tokens}} & \multicolumn{2}{c}{\textbf{Latency (ms)}} \\
\cline{2-3} \cline{4-6}
 & \textbf{Intent} & \textbf{Generation} & \textbf{Intent} & \textbf{Retrieval \&} \\
 & \textbf{Recognition} &  & \textbf{Recognition} & \textbf{Generation} & \\
\hline
CID-GraphRAG & 2136 & 1694 & 2649 & 2501 \\
CID-GraphRAG (optimized) & - & 1694 & 11 & 2501 \\
Intent RAG & 2136 & 1542 & 2649 & 2467 \\
Conversation RAG & - & 1702 & - & 2529 \\
Direct LLM & - & 1203 & - & 2320 \\
\hline
\end{tabular}
\end{table}

While CID-GraphRAG delivers superior retrieval and response quality, it introduces additional computational overhead primarily from LLM-based intent recognition, resulting in higher token consumption and latency compared to baseline approaches. To quantify these computational trade-offs, Table \ref{tab:performance_efficiency} presents a detailed analysis of token consumption and latency across all evaluated approaches. This analysis reveals that intent recognition contributes approximately 
2,136 tokens and 2.6 seconds to the processing pipeline. To explore potential efficiency improvements, we conducted preliminary timing experiments with a BERT-base classifier for intent recognition. This lightweight approach offers 
dual advantages: (1) it completely eliminates the 2,136 input tokens required for intent recognition, reducing per-request cost by approximately 56\%, and (2) it reduces intent recognition latency from 
2.6 seconds to approximately 0.01 seconds (99.6\% reduction), bringing CID-GraphRAG's total latency closer to the Direct LLM baseline. However, the impact of this optimization on retrieval and response quality requires further investigation.

This computational flexibility stems from our framework's modular architecture, which decouples intent recognition from the dual-retrieval mechanism, allowing for domain-specific intent recognition approaches including fine-tuned models, LLM-based classification, or hybrid solutions, depending on specific business requirements and computational constraints.

\subsection {Limitations and Future Work} 

While CID-GraphRAG shows promising results, several limitations and future directions are worth noting:

\textbf{Domain Adaptability}: Our current evaluation focuses on a specific domain of vehicle compliance-related customer service on a focused dataset of 268 conversations. Future work should systematically investigate the framework's transferability to broader multi-turn dialogue scenarios, including complex task-oriented dialogues and cross-domain interactions where conversations span multiple service areas, i.e., well-recognized large-scale datasets such as MultiWOZ.

\textbf{Intent Hierarchy Adaptability:} The current implementation uses a two-level intent hierarchy that balances expressiveness and efficiency for moderate-complexity customer service dialogues. But CID-GraphRAG offers flexibility for different dialogue complexities - simpler dialogues could use a single-layer intent structure, while complex multi-domain conversations might benefit from three or more levels to precisely identify user intents. Future work should systematically evaluate these structural variations to identify optimal configurations for different dialogue complexity levels and domain characteristics.

\textbf{Reinforcement Learning Integration}: The current framework relies on supervised learning from goal-achieved dialogues to construct intent transition graphs. A promising future direction is integrating reinforcement learning techniques to dynamically optimize intent transition probabilities and retrieval weighting parameters based on conversation outcomes. This could enable the system to continuously improve its retrieval strategy.

\section{Conclusion}
To address the challenges in multi-turn customer service dialogue systems, we propose CID-GraphRAG, a novel framework that integrates intent-driven graph structures with semantic similarity retrieval mechanisms. Our approach leverages knowledge graphs to enhance reasoning capabilities and enable more precise retrieval, while systematically incorporating dialogue history to improve contextual understanding and response coherence, thereby significantly boosting answer quality and relevance.

Experimental results demonstrate that CID-GraphRAG significantly outperforms both semantic-only and intent-only baseline approaches across all evaluation criteria, achieving a $57.9\%$ improvement in response quality over conventional RAG methods according to LLM-as-a-Judge head-to-head evaluations. These findings establish CID-GraphRAG as an effective framework for advancing conversational AI in knowledge-intensive domains where both contextual relevance and goal-oriented progression are essential. Future work will extend its applicability to broader multi-turn dialogue scenarios, explore the adaptability of intent hierarchy and integration of reinforcement learning.

\newpage
\bibliography{references}

\begin{thebibliography}{16}
\expandafter\ifx\csname natexlab\endcsname\relax\def\natexlab#1{#1}\fi
\providecommand{\url}[1]{\texttt{#1}}
\providecommand{\href}[2]{#2}
\providecommand{\path}[1]{#1}
\providecommand{\DOIprefix}{doi:}
\providecommand{\ArXivprefix}{arXiv:}
\providecommand{\URLprefix}{URL: }
\providecommand{\Pubmedprefix}{pmid:}
\providecommand{\doi}[1]{\href{http://dx.doi.org/#1}{\path{#1}}}
\providecommand{\Pubmed}[1]{\href{pmid:#1}{\path{#1}}}
\providecommand{\bibinfo}[2]{#2}
\ifx\xfnm\relax \def\xfnm[#1]{\unskip,\space#1}\fi
\bibitem[{Lewis et~al.(2020)Lewis, Perez, Piktus, Petroni, Karpukhin, Goyal, Küttler, Lewis, Yih, Rocktäschel, Riedel, and Kiela}]{Lewis2020}
\bibinfo{author}{P.~Lewis}, \bibinfo{author}{E.~Perez}, \bibinfo{author}{A.~Piktus}, \bibinfo{author}{F.~Petroni}, \bibinfo{author}{V.~Karpukhin}, \bibinfo{author}{N.~Goyal}, \bibinfo{author}{H.~Küttler}, \bibinfo{author}{M.~Lewis}, \bibinfo{author}{W.-t. Yih}, \bibinfo{author}{T.~Rocktäschel}, \bibinfo{author}{S.~Riedel}, \bibinfo{author}{D.~Kiela}, \bibinfo{title}{Retrieval-augmented generation for knowledge-intensive nlp tasks}, \bibinfo{year}{2020}. \URLprefix \url{https://arxiv.org/pdf/2005.11401}. \href{http://arxiv.org/abs/2005.11401}{{\tt arXiv:2005.11401}}.
\bibitem[{Roy et~al.(2024)Roy, Ribeiro, Blloshmi, and Small}]{roy2024learning}
\bibinfo{author}{N.~Roy}, \bibinfo{author}{L.~F. Ribeiro}, \bibinfo{author}{R.~Blloshmi}, \bibinfo{author}{K.~Small},
\newblock \bibinfo{title}{Learning when to retrieve, what to rewrite, and how to respond in conversational qa},
\newblock \bibinfo{journal}{arXiv preprint arXiv:2409.15515}  (\bibinfo{year}{2024}).
\bibitem[{Ye et~al.(2024)Ye, Lei, Yin, Chen, Zhou, and He}]{ye2024boosting}
\bibinfo{author}{L.~Ye}, \bibinfo{author}{Z.~Lei}, \bibinfo{author}{J.~Yin}, \bibinfo{author}{Q.~Chen}, \bibinfo{author}{J.~Zhou}, \bibinfo{author}{L.~He}, \bibinfo{title}{Boosting conversational question answering with fine-grained retrieval-augmentation and self-check}, \bibinfo{year}{2024}. \URLprefix \url{https://arxiv.org/pdf/2403.18243}. \href{http://arxiv.org/abs/2403.18243}{{\tt arXiv:2403.18243}}.
\bibitem[{{Zhu} et~al.(2025){Zhu}, {Huang}, {Wang}, {Ye}, {Chen}, and {Luo}}]{graphrag2025}
\bibinfo{author}{Z.~{Zhu}}, \bibinfo{author}{T.~{Huang}}, \bibinfo{author}{K.~{Wang}}, \bibinfo{author}{J.~{Ye}}, \bibinfo{author}{X.~{Chen}}, \bibinfo{author}{S.~{Luo}}, \bibinfo{title}{Graph-based approaches and functionalities in retrieval-augmented generation: A comprehensive survey}, \bibinfo{year}{2025}. \URLprefix \url{https://arxiv.org/html/2504.10499v1}, \bibinfo{note}{accessed: 2025-06-17}.
\bibitem[{Edge et~al.(2024)Edge, Trinh, Cheng, Bradley, Chao, Mody, Truitt, Metropolitansky, Ness, and Larson}]{graphrag2024}
\bibinfo{author}{D.~Edge}, \bibinfo{author}{H.~Trinh}, \bibinfo{author}{N.~Cheng}, \bibinfo{author}{J.~Bradley}, \bibinfo{author}{A.~Chao}, \bibinfo{author}{A.~Mody}, \bibinfo{author}{S.~Truitt}, \bibinfo{author}{D.~Metropolitansky}, \bibinfo{author}{R.~O. Ness}, \bibinfo{author}{J.~Larson}, \bibinfo{title}{From local to global: A graphrag approach to query-focused summarization}, \bibinfo{year}{2024}. \URLprefix \url{https://arxiv.org/html/2404.16130}, \bibinfo{note}{accessed: 2025-06-17}.
\bibitem[{Zhang et~al.(2025{\natexlab{a}})Zhang, Tan, Yang, Deng, and Wang}]{zhang2025credible}
\bibinfo{author}{C.~Zhang}, \bibinfo{author}{Z.~Tan}, \bibinfo{author}{X.~Yang}, \bibinfo{author}{W.~Deng}, \bibinfo{author}{W.~Wang}, \bibinfo{title}{Credible plan-driven rag method for multi-hop question answering}, \bibinfo{year}{2025}{\natexlab{a}}. \URLprefix \url{https://arxiv.org/html/2504.16787}. \href{http://arxiv.org/abs/2504.16787}{{\tt arXiv:2504.16787}}.
\bibitem[{Zhang et~al.(2025{\natexlab{b}})Zhang, Zhu, Ming, Jin, Chai, Yang, Tian, Fan, and Chen}]{zhang2025dhrag}
\bibinfo{author}{F.~Zhang}, \bibinfo{author}{D.~Zhu}, \bibinfo{author}{J.~Ming}, \bibinfo{author}{Y.~Jin}, \bibinfo{author}{D.~Chai}, \bibinfo{author}{L.~Yang}, \bibinfo{author}{H.~Tian}, \bibinfo{author}{Z.~Fan}, \bibinfo{author}{K.~Chen}, \bibinfo{title}{Dh-rag: A dynamic historical context-powered retrieval-augmented generation method for multi-turn dialogue}, \bibinfo{year}{2025}{\natexlab{b}}. \URLprefix \url{https://arxiv.org/pdf/2502.13847}. \href{http://arxiv.org/abs/2502.13847}{{\tt arXiv:2502.13847}}.
\bibitem[{Asai et~al.(2023)Asai, Wu, Wang, Sil, and Hajishirzi}]{self-rag}
\bibinfo{author}{A.~Asai}, \bibinfo{author}{Z.~Wu}, \bibinfo{author}{Y.~Wang}, \bibinfo{author}{A.~Sil}, \bibinfo{author}{H.~Hajishirzi}, \bibinfo{title}{Self-rag: Learning to retrieve, generate, and critique through self-reflection}, \bibinfo{year}{2023}. \URLprefix \url{https://arxiv.org/pdf/2310.11511}.
\bibitem[{Katsis et~al.(2025)Katsis, Rosenthal, Fadnis, Gunasekara, Lee, Popa, Shah, Zhu, Contractor, and Danilevsky}]{katsis2025mtrag}
\bibinfo{author}{Y.~Katsis}, \bibinfo{author}{S.~Rosenthal}, \bibinfo{author}{K.~Fadnis}, \bibinfo{author}{C.~Gunasekara}, \bibinfo{author}{Y.-S. Lee}, \bibinfo{author}{L.~Popa}, \bibinfo{author}{V.~Shah}, \bibinfo{author}{H.~Zhu}, \bibinfo{author}{D.~Contractor}, \bibinfo{author}{M.~Danilevsky}, \bibinfo{title}{Mtrag: A multi-turn conversational benchmark for evaluating retrieval-augmented generation systems}, \bibinfo{year}{2025}. \URLprefix \url{https://arxiv.org/abs/2501.03468}. \href{http://arxiv.org/abs/2501.03468}{{\tt arXiv:2501.03468}}.
\bibitem[{Xu et~al.(2024)Xu, Cruz, Guevara, Wang, Deshpande, Wang, and Li}]{xu2024retrieval}
\bibinfo{author}{Z.~Xu}, \bibinfo{author}{M.~J. Cruz}, \bibinfo{author}{M.~Guevara}, \bibinfo{author}{T.~Wang}, \bibinfo{author}{M.~Deshpande}, \bibinfo{author}{X.~Wang}, \bibinfo{author}{Z.~Li}, \bibinfo{title}{Retrieval-augmented generation with knowledge graphs for customer service question answering}, \bibinfo{year}{2024}. \URLprefix \url{https://arxiv.org/abs/2404.17723}. \href{http://arxiv.org/abs/2404.17723}{{\tt arXiv:2404.17723}}.
\bibitem[{Agrawal et~al.(2024)Agrawal, Gummuluri, and Spera}]{agrawal2024beyond}
\bibinfo{author}{G.~Agrawal}, \bibinfo{author}{S.~Gummuluri}, \bibinfo{author}{C.~Spera}, \bibinfo{title}{Beyond-rag: Question identification and answer generation in real-time conversations}, \bibinfo{year}{2024}. \URLprefix \url{https://arxiv.org/html/2410.10136v1}. \href{http://arxiv.org/abs/2410.10136}{{\tt arXiv:2410.10136}}, \bibinfo{note}{arXiv:2410.10136v1}.
\bibitem[{Linders and Tomczak(2025)}]{linders2025knowledge}
\bibinfo{author}{J.~Linders}, \bibinfo{author}{J.~M. Tomczak}, \bibinfo{title}{Knowledge graph-extended retrieval augmented generation for question answering}, \bibinfo{year}{2025}. \URLprefix \url{https://arxiv.org/pdf/2504.08893}. \href{http://arxiv.org/abs/2504.08893}{{\tt arXiv:2504.08893}}.
\bibitem[{Xu et~al.(2025)Xu, Shi, Zhuang, Yu, Ho, Wang, and Yang}]{collab-rag}
\bibinfo{author}{R.~Xu}, \bibinfo{author}{W.~Shi}, \bibinfo{author}{Y.~Zhuang}, \bibinfo{author}{Y.~Yu}, \bibinfo{author}{J.~C. Ho}, \bibinfo{author}{H.~Wang}, \bibinfo{author}{C.~Yang}, \bibinfo{title}{Collab-rag: Boosting retrieval-augmented generation for complex question answering via white-box and black-box llm collaboration}, \bibinfo{year}{2025}. \URLprefix \url{https://arxiv.org/pdf/2504.04915}.
\bibitem[{Kuo et~al.(2024)Kuo, Liao, Hsieh, Chang, Hsu, and Shiu}]{kuo2024radbench}
\bibinfo{author}{T.-L. Kuo}, \bibinfo{author}{F.-T. Liao}, \bibinfo{author}{M.-W. Hsieh}, \bibinfo{author}{F.-C. Chang}, \bibinfo{author}{P.-C. Hsu}, \bibinfo{author}{D.-S. Shiu}, \bibinfo{title}{Rad-bench: Evaluating large language models' capabilities in retrieval augmented dialogues}, \bibinfo{year}{2024}. \URLprefix \url{https://arxiv.org/abs/2409.12558}. \href{http://arxiv.org/abs/2409.12558}{{\tt arXiv:2409.12558}}.
\bibitem[{Liu et~al.(2024)Liu, Ping, Roy, Xu, Lee, Shoeybi, and Catanzaro}]{chatqa}
\bibinfo{author}{Z.~Liu}, \bibinfo{author}{W.~Ping}, \bibinfo{author}{R.~Roy}, \bibinfo{author}{P.~Xu}, \bibinfo{author}{C.~Lee}, \bibinfo{author}{M.~Shoeybi}, \bibinfo{author}{B.~Catanzaro}, \bibinfo{title}{Chatqa: Surpassing gpt-4 on conversational qa and rag}, \bibinfo{year}{2024}. \URLprefix \url{https://openreview.net/pdf?id=bkUvKPKafQ}.
\bibitem[{{AWS Sonnet Documentation}(2023)}]{sonnet_aws}
\bibinfo{author}{{AWS Sonnet Documentation}}, \bibinfo{title}{Sonnet 3.7 on amazon bedrock}, \bibinfo{howpublished}{\url{https://docs.aws.amazon.com/bedrock/latest/userguide/model-evaluation-type-judge-prompt-claude-sonnet37.html}}, \bibinfo{year}{2023}. \bibinfo{note}{Accessed: 2024}.

\end{thebibliography}

\section*{Declaration on Generative AI}
During the preparation of this work, the authors used Claude 4.5 Sonnet in order to improve writing style. After using this tool, the authors reviewed and edited the content as needed and takes full responsibility for the publication's content.

\newpage
\appendix
\section{Intent Recognition Examples}
\label{app:intent_recognition}
Our dual-layer intent framework is essential for disambiguating utterances with identical wording but different contextual meanings. Consider these examples from a vehicle compliance topic:

\begin{tcolorbox}[colback=gray!5, colframe=gray!50]
\fontsize{8pt}{11pt}\selectfont 
\textbf{Scenario 1. Refusing a proposed appointment time}

\textbf{Assistant}: I see your vehicle compliance sticker expired last week. We have an opening tomorrow at 2:00 PM to replace it and bring your vehicle back into compliance.

\textbf{User}: That doesn't work for me.\\

\textbf{Primary intent ($I^{\text{user}}_1$):} Appointment Schedule

\textbf{Secondary intent ($I^{\text{user}}_2$):} Timing Rejection

\textbf{Appropriate assistant response:} No problem. We have several other time slots available this week. Would Thursday morning at 9:30 AM or Friday afternoon at 3:15 PM work better for your schedule?\\

\vspace{1em}
\textbf{Scenario 2. Refusing a service center location}

\textbf{Assistant}: Your vehicle's compliance sticker needs replacement to avoid penalties. You can bring your vehicle to our main service center on XYZ Boulevard for the replacement.

\textbf{User}: That doesn't work for me.\\

\textbf{Primary intent ($I^{\text{user}}_1$):} Appointment Schedule

\textbf{Secondary intent ($I^{\text{user}}_2$):} Location Rejection

\textbf{Appropriate assistant response:} I understand that location might be inconvenient. We also have service centers in A Mall and B Plaza.
\end{tcolorbox}

While these user utterances share the same primary intent, their secondary intents differ substantially, requiring distinct system responses.

\section{Intent Fallback Mechanism for Novel Intent Combinations}
\label{app:fallback}
As discussed in Section~\ref{sec:dual_retrieval}, a critical challenge in intent-based retrieval is handling previously unseen intent combinations. Our analysis of the validation set revealed that 58 dialogue turns (39\% of total) had no exact intent pair matches in the intent graph, making a robust fallback mechanism essential.

\subsection{Semantic Matching Algorithm}
When no matching intent pattern exists for a given intent pair $P(I^{\text{assistant}}_2, I^{\text{user}}_2)$, our system implements the following fallback procedure:

\begin{enumerate}
    \item \textbf{Intent Pair Embedding:} The system embeds the current intent pair using the BGE-M3 model:
    \begin{equation}
        e_{\text{curr}} = \text{BGE-M3}(P(I^{\text{assistant}}_2, I^{\text{user}}_2))
    \end{equation}

    \item \textbf{Candidate Filtering:} The system filters candidate intent pairs to those sharing identical primary intents with the current pair:
    \begin{equation}
    \begin{split}
        \text{Candidates} = \{P_i \mid  P_i.I^{\text{assistant}}_1 == P.I^{\text{assistant}}_1 \text{ AND } P_i.I^{\text{user}}_1 == P.I^{\text{user}}_1\}
    \end{split}
    \end{equation}

    \item \textbf{Similarity Ranking:} For each candidate intent pair, we compute cosine similarity:
    \begin{equation}
        \text{sim}(P, P_i) = \text{cos}(e_{\text{curr}}, \text{BGE-M3}(P_i))
    \end{equation}

    \item \textbf{Threshold-Based Selection:} The system selects the highest-scoring candidate over threshold $\tau$:
    \begin{equation}
    \begin{split}
         P_{\text{selected}} = \underset{i}{\text{argmax}}(\text{sim}(P, P_i)) \text{ if } \max(\text{sim}(P, P_i)) > \tau
    \end{split}
    \end{equation}
\end{enumerate}

\subsection{Effectiveness Evaluation}

To evaluate our fallback mechanism, we isolated 58 dialogue turns with no exact intent matches in the validation set and compared two approaches:
\begin{enumerate}
    \item Exact matching only (failing to retrieve any examples)
    \item Semantic matching with $\tau = 0.8$
\end{enumerate}

\begin{figure}[t]
\centering
\includegraphics[width=0.9\linewidth]{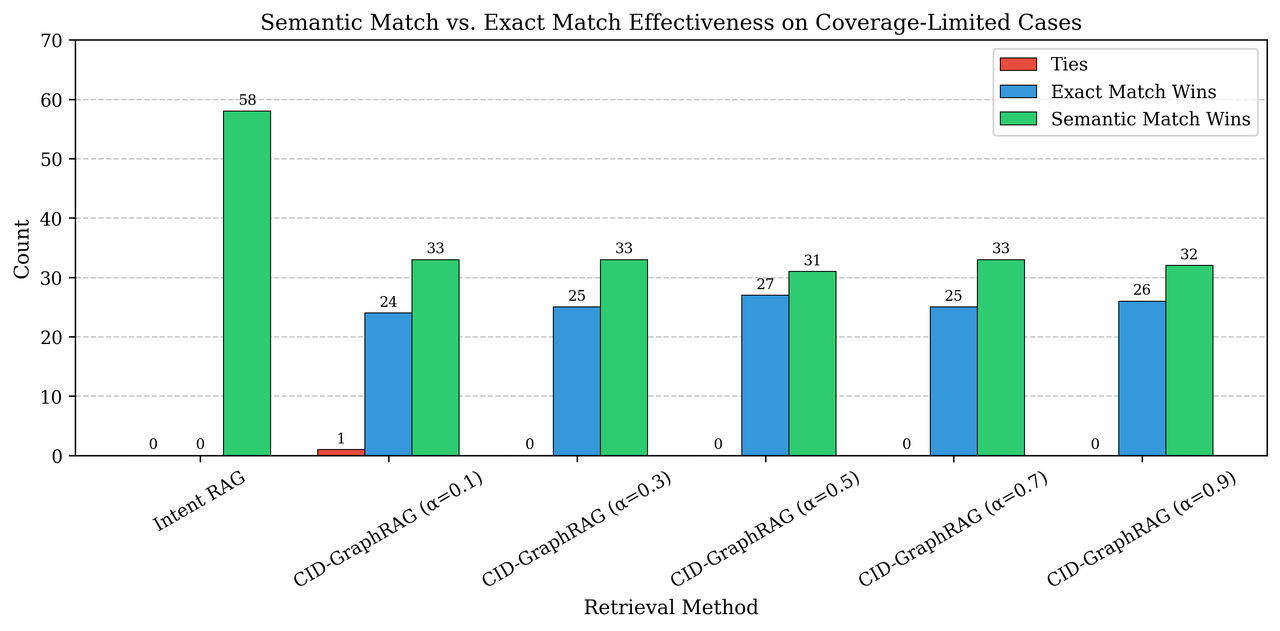}
\caption{Comparison between semantic matching and exact matching on 58 cases. Semantic matching outperforms exact matching across all weight configurations in CID-GraphRAG.}
\label{fig:semantic_match_comparison}
\end{figure}

We use LLM-as-a-Judge to evaluate the generated response quality. Figure~\ref{fig:semantic_match_comparison} shows that semantic matching significantly outperforms exact matching across all weight configurations, confirming the importance of this fallback mechanism in practical deployments.

\section{Conversation-Based Retrieval Implementation Details}
\label{app:semantic_match}
As described in Section~\ref{sec:dual_retrieval}, the conversation-based retrieval path identifies semantically similar historical dialogues to enhance response generation. Here we provide the technical implementation details of this process.

Our system processes dialogue context through the following pipeline:

\begin{enumerate}
    \item \textbf{Context Embedding:} We encode both current and historical dialogues using the BGE-M3 embedding model:
    \begin{align}
        e_{\text{curr}} &= \text{BGE-M3}(D_{\text{curr}}) \\
        e_{\text{hist}} &= \text{BGE-M3}(D_{\text{hist}})
    \end{align}
    
    \item \textbf{Similarity Computation:} We compute the semantic similarity between dialogue contexts using cosine similarity:
    \begin{equation}
        \text{sim}(D_{\text{hist}}, D_{\text{curr}}) = \cos(e_{\text{hist}}, e_{\text{curr}})
    \end{equation}
\end{enumerate}

For each candidate assistant intent $I^{\text{candidate}}_2$ identified through the intent-based retrieval path, we retrieve the top historical dialogues with the highest similarity scores. These retrieved dialogues complementing the structural guidance provided by the intent-based retrieval path.

\section{LLM-as-a-Judge Evaluation Prompt}
\label{judge_prompt}
For reproducibility, we provide the complete prompt template used for LLM-as-a-Judge evaluation:
\begin{tcolorbox}[colback=gray!5, colframe=gray!50]
\fontsize{8pt}{11pt}\selectfont 

You are a professional freight platform dialogue evaluation expert. Please evaluate the responses generated by the following four RAG systems and determine which one better fulfills the role requirements of a freight platform representative.\\

\# Conversation Background\\
A freight platform representative is contacting a user to inform a compliance issue and to guide the user to resolve the issue.\\

\# Conversation History\\
\{conversation\_history\}\\

\# User's Latest Response\\
\{user\_response\}\\

\# Ground Truth Response \\
\{ground\_truth\_response\}\\

\# System A Generated Response\\
\{system\_a\_response\}\\
\# System B Generated Response\\
\{system\_b\_response\}\\
\# System C Generated Response\\
\{system\_c\_response\}\\
\# System D Generated Response\\
\{system\_d\_response\}\\

Please evaluate the generated responses based on the following specific criteria for freight platform representatives:\\
1. Relevance: Whether the response delivers directly relevant information in a concise, straightforward manner\\
2. Style Consistency: Whether it matches conversational style of the representative and appropriately uses friendly terms\\
3. Helpfulness: Whether it effectively answers the user's questions or guides them to resolve the issue\\
4. Contextual Appropriateness: Whether there is obvious repetition or similarity with the representative's responses in the historical dialogue\\
5. Professional Standards: Whether the response content meets the professional standards of the freight industry representative\\
Please provide your evaluation in the following format:\\
\{\\
  "final\_rating": \{\\
      "System A": 1-10 points,\\
      "System B": 1-10 points,\\
      "System C": 1-10 points,\\
      "System D": 1-10 points,\\
      "Analysis": "Brief explanation of the final rating reasons, analyzing the overall advantages and disadvantages of the four systems"\\
    \}\\
  "detailed\_ratings": \{\\
    "Relevance": \{\\
      "System A": 1-10 points,\\
      "System B": 1-10 points,\\
      "System C": 1-10 points,\\
      "System D": 1-10 points,\\
      "Analysis": "Brief analysis of the four systems' performance in this dimension"\\
    \},\\
    "Style Consistency": \{\\
    ...
    \},\\
    ...\\
  \}\\
\}\\

\end{tcolorbox}

\section{Dataset Example}
\label{app:dataset_example}
Below is a representative example from our customer service dialogue dataset, demonstrating the multi-turn nature and task-oriented characteristics of the conversations:

\begin{tcolorbox}[colback=gray!5, colframe=gray!50]
\fontsize{8pt}{11pt}\selectfont 
Assistant: Good morning, I'm calling from Company X about an urgent matter regarding vehicle ABC-123. Our system flagged your vehicle for a damaged compliance sticker, which could result in penalties. Have you had a chance to address this?

User: Oh, I didn't realize it had damaged. What do I need to do?

Assistant: I understand. To avoid potential fines, you'll need to replace your sticker as soon as possible. Would tomorrow work for you?

User: I'm quite busy tomorrow.

Assistant: No problem. We have several other time slots available this week. Would Thursday morning at 9:30 AM or Friday afternoon at 3:15 PM work better for your schedule?

User: Thursday morning works for me.

Assistant: Perfect! I've scheduled your appointment for Thursday at 9:30 AM at our main service center on 123 Transport Avenue. Please bring your vehicle and confirmation email which I'll send shortly. Looking forward to seeing you!

User: Sounds good. See you soon.
\end{tcolorbox}

This example demonstrates a successful goal-achieved dialogue where the assistant effectively addresses the user's vehicle sticker compliance issue, collects necessary information, provides clear guidance, and schedules an appointment, all while maintaining a professional and helpful tone.

\end{document}